\documentclass{article}

\usepackage{PRIMEarxiv}

\usepackage[utf8]{inputenc} 
\usepackage[T1]{fontenc}    
\usepackage{hyperref}       
\usepackage{url}            
\usepackage{booktabs}       
\usepackage{amsfonts}       
\usepackage{nicefrac}       
\usepackage{microtype}      
\usepackage{lipsum}
\usepackage{fancyhdr}       
\usepackage{graphicx}       

\usepackage{xcolor}
\usepackage{todonotes}
\usepackage{amsmath}
\usepackage{subcaption}
\usepackage{caption} 
\usepackage{amsfonts}

\usepackage{listings}
\usepackage{xcolor}       
\usepackage{multirow}
\usepackage{tcolorbox}
\usepackage{tikz}
\usetikzlibrary{decorations.pathreplacing}

\colorlet{lightyellow}{yellow!40}

\makeatletter
{\small 
\xdef\f@size@small{\f@size}
\xdef\f@baselineskip@small{\f@baselineskip}
\normalsize 
\xdef\f@size@normalsize{\f@size}
\xdef\f@baselineskip@normalsize{\f@baselineskip}
}
\newcommand{\smalltonormalsize}{%
  \fontsize
    {\fpeval{(\f@size@small+\f@size@normalsize)/2}}
    {\fpeval{(\f@baselineskip@small+\f@baselineskip@normalsize)/2}}%
  \selectfont
}
\makeatother

\title{Many-Shot Regurgitation (MSR) Prompting}

\author{
  Shashank Sonkar \\
  Rice University \\
  Houston, TX \\
  \texttt{ss164@rice.edu} \\
  \And
  Richard G. Baraniuk \\
  Rice University \\
  Houston, TX \\
  \texttt{richb@rice.edu} \\
}

\begin{document}
\maketitle

\begin{abstract}
We introduce Many-Shot Regurgitation (MSR) prompting, a new black-box membership inference attack framework for examining verbatim content reproduction in large language models (LLMs). MSR prompting involves dividing the input text into multiple segments and creating a single prompt that includes a series of faux conversation rounds between a user and a language model to elicit verbatim regurgitation. We apply MSR prompting to diverse text sources, including Wikipedia articles and open educational resources (OER) textbooks, which provide high-quality, factual content and are continuously updated over time. For each source, we curate two dataset types: one that LLMs were likely exposed to during training ($D_{\rm pre}$) and another consisting of documents published after the models' training cutoff dates ($D_{\rm post}$). To quantify the occurrence of verbatim matches, we employ the Longest Common Substring algorithm and count the frequency of matches at different length thresholds. We then use statistical measures such as Cliff's delta, Kolmogorov-Smirnov (KS) distance, and Kruskal-Wallis H test to determine whether the distribution of verbatim matches differs significantly between $D_{\rm pre}$ and $D_{\rm post}$. Our findings reveal a striking difference in the distribution of verbatim matches between $D_{\rm pre}$ and $D_{\rm post}$, with the frequency of verbatim reproduction being significantly higher when LLMs (e.g. GPT models and LLaMAs) are prompted with text from datasets they were likely trained on. For instance, when using GPT-3.5 on Wikipedia articles, we observe a substantial effect size (Cliff's delta $= -0.984$) and a large KS distance ($0.875$) between the distributions of $D_{\rm pre}$ and $D_{\rm post}$. Our results provide compelling evidence that LLMs are more prone to reproducing verbatim content when the input text is likely sourced from their training data.

\end{abstract}

\section{Introduction}
Large language models (LLMs) have revolutionized the field of natural language processing (NLP) with their remarkable ability to generate human-like text across a wide range of domains. These models, such as GPT \cite{gpt4_technical_report}, Gemini \cite{team2023gemini}, and LLaMA \cite{touvron2023llama,llama3modelcard}, are trained on vast amounts of diverse text data, enabling them to produce coherent and contextually relevant responses to given prompts. 
Important emerging questions relevant to artificial intelligence (AI) alignment \cite{wikipediaAIAlignment,openai_alignment} revolve around the how and where a model’s training data was obtained.
In this paper, we study the following specific question that is relevant to studying training data provenance: to what extent do these models reproduce verbatim content from their training data? 
We develop the Many-Shot Regurgitation (MSR) prompting technique, a new approach designed to efficiently investigate this phenomenon.
Code is available on github\footnote{\url{https://github.com/luffycodes/Many-Shot-Regurgitation-MIA}}.

MSR represents a new framework for membership inference attack (MIA) against language models. MIAs aim to determine whether a given input was part of a model's training data and have been studied extensively in both computer vision \cite{carlini2022membership,carlini2022privacy} and natural language processing domains \cite{song2020information,mattern2023membership,mireshghallah2022empirical}. However, most prior MIA research on language models has focused on the fine-tuning stage \cite{fu2023practical,mireshghallah2022empirical,mattern2023membership}and assumes access to model internals such as logits, token probabilities, or loss values \cite{shokri2017membership,carlini2021extracting,duan2024membership}. In contrast, MSR operates in a black-box setting, requiring only the ability to prompt the model and observe its outputs. This makes MSR applicable to studying real-world deployed models like GPT-3.5 \cite{schulman2022chatgpt} and GPT-4 \cite{gpt4_technical_report,openai2023gpt4}, where access to model internals is no longer available due to API changes.

The MSR prompting technique, inspired by the many-shot jailbreaking approach \cite{chao2023jailbreaking,anilmany}, involves dividing the input text into multiple segments and creating a single prompt that includes a series of faux conversation rounds between a user and a language model designed to elicit verbatim regurgitation. The input text is segmented into multiple parts, denoted as $T_1$, $T_2$, ..., $T_n$, where $n$ is the total number of segments. A single prompt is then constructed, creating a simulated conversation sequence between a user and a language model. Throughout these rounds of conversation, the user supplies the odd-numbered segments ($T_1$, $T_3$, ..., $T_{n-1}$), while the even-numbered segments ($T_2$, $T_4$, ..., $T_{n-2}$) are presented as if they were generated by the language model. In the final round, the user presents the penultimate segment ($T_{n-1}$), prompting the language model to generate the concluding segment ($T_n'$). It is worth highlighting that the segments presented as responses from the language model are not genuine outputs but are integrated into the prompt to mimic a conversation. The sole genuine output from the LLM is the final segment ($T_n'$).

While quantifying the occurrence of verbatim matches between the generated text ($T_n'$) and the original segment ($T_n$) is important, it doesn't provide a complete understanding of verbatim reproduction. Thus, we also explore whether the distribution of these verbatim matches is statistically different from matches in documents that the LLMs are unlikely to have been trained on. To achieve this, we curate two types of dataset sources: $D_{\rm pre}$, which includes documents that the LLMs were likely exposed to during training, and $D_{\rm post}$, which consists of documents that were published after the LLMs' training cutoff dates. Specifically, we utilize Wikipedia articles and Open Educational Resource (OER) textbooks as they provide high-quality, factual content and are continuously updated over time.

Through experiments employing the MSR prompting technique on both $D_{\rm pre}$ and $D_{\rm post}$ datasets, we aim to compare the distributions of verbatim matches. We employ statistical tests such as Cliff's Delta \cite{cliffdelta}, Kolmogorov-Smirnov (KS) Distance \cite{KolmogorovSmirnovDist}, and the Kruskal-Wallis H Test \cite{kruskalWallisTest} to evaluate the statistical significance of the differences in verbatim reproduction between the two dataset types. Our experiments are performed using three state-of-the-art LLMs: GPT-3.5 \cite{schulman2022chatgpt}, GPT-4 \cite{gpt4_technical_report}, and LLaMA-3 \cite{llama3modelcard}.

Our experiments reveal that verbatim reproduction occurs in text generated by different LLMs, with the extent of reproduction varying across dataset sources and language models. We find that the frequency of verbatim matches is generally higher for $D_{\rm pre}$ compared to $D_{\rm post}$, suggesting that the availability of the dataset influences the likelihood of verbatim reproduction. For example, using GPT-3.5 on Wikipedia articles, we observe a Cliff's delta of $-0.984$ and a Kolmogorov-Smirnov distance of $0.875$, indicating a significant difference in the distribution of verbatim matches between $D_{\rm pre}$ and $D_{\rm post}$. Furthermore, we investigate the impact of factors such as the number of shots in MSR prompting and the temperature settings of LLMs on the occurrence of verbatim reproduction. We find that increasing the number of shots generally leads to higher frequencies of verbatim matches, with 6 shots yielding the best results. Additionally, lower temperature settings favor more deterministic and less diverse outputs, leading to a higher likelihood of verbatim regurgitation.

The main contributions of this paper are as follows:
\begin{itemize}
    \item We introduce the Many-Shot Regurgitation (MSR) prompting technique, a novel approach to investigate verbatim reproduction in LLMs across diverse domains and dataset sources.
    \item We propose a methodology that not only quantifies the occurrence of verbatim matches but also determines whether the distribution of these matches is statistically different from matches in documents that the LLMs are unlikely to have been trained on, using statistical measures such as Cliff's delta, Kolmogorov-Smirnov distance, and the Kruskal-Wallis H test.
    \item We conduct extensive experiments using the MSR prompting technique on carefully curated dataset sources ($D_{\rm pre}$ and $D_{\rm post}$), including Wikipedia articles and OER textbooks, and three state-of-the-art LLMs (GPT-3.5, GPT-4, and LLAMA) to quantify the extent of verbatim reproduction and analyze the influence of dataset availability on the occurrence of verbatim reproduction.
    \item We investigate the impact of factors such as the number of shots in MSR prompting and the temperature settings of LLMs on the likelihood of verbatim reproduction.
\end{itemize}

The remainder of this paper is organized as follows: Section 2 provides an overview of related work on verbatim reproduction and memorization in LLMs. Section 3 describes the proposed MSR prompting technique and the dataset sources used in our experiments. Section 4 presents the experimental setup, results, and analysis of our findings. Finally, Section 5 concludes the paper and discusses the implications of our work for the development and usage of LLMs in various domains.

\section{Related Work}
Membership Inference Attacks (MIAs) aim to determine whether a given data sample was part of a model's training data \cite{shokri2017membership,carlini2021extracting}. MIAs have been extensively studied in both computer vision \cite{carlini2022membership,carlini2022privacy,zarifzadeh2023low} and natural language processing domains \cite{song2020information,mattern2023membership,mireshghallah2022empirical,fu2023practical}. These attacks have implications for quantifying privacy risks \cite{mireshghallah2022quantifying}, measuring memorization \cite{carlini2022quantifying}, auditing models \cite{steinke2024privacy,yao2024machine}, and detecting issues like test set contamination \cite{oren2023proving} and unauthorized use of copyrighted content \cite{meeus2023did,duarte2024cop}. In the language modeling domain, most MIA research focuses on the fine-tuning stage \cite{fu2023practical,mireshghallah2022empirical,mattern2023membership} rather than pretraining. \cite{carlini2021extracting} demonstrated that language models can memorize verbatim content from training data, but their method requires manually inspecting model outputs. Subsequent work has explored more automated approaches. \cite{song2020information} studied membership inference on embedding models, showing that they can leak sensitive attributes. \cite{pan2020privacy} proposed a method for auditing embedding models' training data using gradient updates. For auto-regressive language models, \cite{carlini2021extracting} introduced the exposure metric to measure data leakage but note that Monte Carlo estimation of this metric is challenging for large models. \cite{kandpal2022deduplicating} showed that deduplicating training data can mitigate privacy risks, but doing so for large web-scale corpora is difficult. Membership inference against large pre-trained language models poses unique challenges compared to conventional MIA settings, such as limited access to training data and model scale. Assumptions like having shadow models trained on the same data distribution often do not hold, as pretraining corpora tend to be proprietary \cite{shi2023detecting,duan2024membership}. The sheer scale of pretraining, involving massive datasets and relatively few epochs, makes MIAs inherently more difficult.

Recent work has started exploring pretraining data detection for large language models. \cite{shi2023detecting} proposed the Min-K\% method which uses the average log probability of low-probability tokens as a score for pretraining data detection. However, this approach relies on heuristics and access to token probabilities which are no longer available for models like GPT-3. \cite{duan2024membership} constructed the MIMIR benchmark and analyzed the challenges of pretraining data detection, noting the need for better methods. Most prior MIA methods assume some form of access to a model's internals such as logits, gradients, loss values or token probabilities \cite{shokri2017membership,carlini2021extracting,mattern2023membership,steinke2024privacy}. However, this poses challenges for studying deployed language models in real-world settings. For example, the GPT-3 API previously provided access to logits and token probabilities but this functionality has been deprecated \cite{shi2023detecting,duan2024membership}, precluding the use of methods that rely on these values. Access to gradients or internal model activations is even more restricted. Therefore, there is a need for MIA frameworks that can operate under black-box settings with only API-level access. 

Our work addresses this gap by introducing the Many-Shot Regurgitation (MSR) technique, a black-box MIA framework that requires no access to model internals. MSR exploits LLMs' conversational and continuation abilities to elicit verbatim regurgitation, which can then be precisely quantified and statistically analyzed. This differentiates MSR from prior approaches that rely on manual inspection \cite{carlini2021extracting}, heuristics based on token probabilities \cite{shi2023detecting}, or access to the full training dataset.

While most MIA research focuses on detecting individual training examples, our work takes a distributional approach to assess a model's overall propensity for verbatim regurgitation. By comparing the frequency distributions of verbatim matches between datasets that were likely seen during training ($D_{\rm pre}$) and those that were not ($D_{\rm post}$), we can quantify the influence of dataset availability on regurgitation behavior. 
This comparative analysis is particularly valuable in settings where verbatim reproduction is rare but still concerning, such as identifying the presence of copyrighted articles or book chapters in training data.
Even if only a small fraction of a document's content is reproduced verbatim, it can serve as evidence of that document's inclusion in the training corpus.


\section{Methodology}
In this study, we introduce the Many-Shot Regurgitation (MSR) prompting technique to investigate the extent of verbatim regurgitation in text generated by language models. The primary objective of our methodology is to quantify the occurrence of verbatim matches between the generated text and the source material, and to determine if the distribution of these matches differs significantly between datasets that the language models were likely trained on and those that were released after the models' training cutoff dates.

\subsection{Dataset Curation}
To achieve our research goals, we carefully curate two types of dataset sources: Wikipedia articles and Open Educational Resource (OER) textbooks. These sources are chosen for their high-quality, factual content and the continuous addition of new material over time.

Wikipedia, being one of the largest and most comprehensive online encyclopedias, covers a wide range of topics and is frequently updated with new articles and information. By selecting Wikipedia articles written in different time periods, we can effectively compare the verbatim regurgitation behavior of language models on articles that were likely included in their training data and those that were added after the models' training cutoff dates.

Similarly, OER textbooks provide a rich source of educational content across various subjects. The release of new OER textbooks and the updating of existing ones allow us to investigate the extent of verbatim regurgitation in language models when prompted with textbook content that was available during their training and content that was published later.

For each dataset source, we select a subset of documents that were likely included in the language models' training data ($D_{\rm pre}$) and a subset of documents that were released after the models' training cutoff dates ($D_{\rm post}$). By creating this controlled setting, we aim to investigate the influence of dataset age and availability on the extent of verbatim regurgitation and draw insights into the models' ability to generate novel content.

\subsection{Many-Shot Regurgitation (MSR) Prompting}
The MSR prompting technique involves dividing the source text into multiple segments (shots) and creating a single prompt with a series of alternating turns between a user and the language model. The key steps of the MSR prompting technique are as follows:

\begin{enumerate}
    \item Text Segmentation: The source text is divided into an even number of segments, denoted as $T_1$, $T_2$, ..., $T_n$, where $n$ is the total number of segments.
    \item Prompt Construction: A single prompt is created by concatenating the segments in an alternating manner, simulating a conversation between a user and the language model. The user provides the odd-numbered segments ($T_1$, $T_3$, ..., $T_{n-1}$), and the language model is prompted to generate the even-numbered segments ($T_2$, $T_4$, ..., $T_n$).
    \begin{tcolorbox}
    \begin{minipage}{\dimexpr\textwidth-7cm}
    System: complete the paragraph \\
    User: $T_1$ \\
    LLM: $T_2$ \\
    User: $T_3$ \\
    LLM: $T_4$ \\
    ... \\
    User: $T_{n-1}$ \\
    LLM: (generates $T_n'$)
    \end{minipage}
    \hfill
    \begin{tikzpicture}[remember picture, overlay]
    \draw [decoration={brace,amplitude=0.7em},decorate,ultra thick]
    (-9.5,5.3em) -- (-9.5,-2.7em)
    node [midway,right=0.8em,align=center,text width=4cm] {single prompt with alternating turns};
    \end{tikzpicture}
    \end{tcolorbox}
    It's important to note that the parts labeled as LLM's responses ($T_2$, $T_4$, ..., $T_{n-2}$) are not actually generated by the model but are included in the prompt to create the illusion that LLM has generated those parts. Only the last part ($T_n'$) is the genuine output from the LLM.
    \item Text Generation: The language model is tasked with generating the final segment $T_n$ based on the constructed prompt. The generated segment is denoted as $T_n'$.
\end{enumerate}

By presenting the source text in this format, the MSR prompting technique aims to examine how language models respond when prompted to complete a passage that closely resembles the original source material.

\subsection{Verbatim Match Analysis}
To quantify the occurrence of verbatim regurgitation, we analyze the generated text ($T_n'$) and compare it with the corresponding original segment ($T_n$) using the Longest Common Substring (LCS) algorithm. The LCS algorithm identifies the longest contiguous substring that is present in both the generated and original segments.

We define a verbatim match as a substring of a specified minimum length (e.g., $\ell_{\rm min}$ to $\ell_{\rm max}$ words) that is identical between the generated and original segments. Formally, let $M_k$ be the set of all substring matches of length $k$ between $T_n'$ and $T_n$. The frequency of exact verbatim matches of length $k$ is defined as:

\begin{equation}
f_k = |M_k|
\end{equation}

For each language model and dataset source combination, we obtain an array of frequencies $F = [f_{\ell_{\rm min}}, f_{\ell_{\rm min}+1}, ..., f_{\ell_{\rm max}}]$ corresponding to verbatim match length thresholds from $\ell_{\rm min}$ to $\ell_{\rm max}$ words.

By counting the frequency of verbatim matches at different length thresholds, we obtain a distribution of verbatim match frequencies for each combination of language model and dataset source.

\subsection{Statistical Analysis}
To determine if the distribution of verbatim matches differs significantly between $D_{\rm pre}$ and $D_{\rm post}$, we perform statistical analysis on the arrays of frequencies $F$ obtained for each language model and dataset source combination. We employ several statistical measures:

\begin{itemize}
    \item Cliff's delta \cite{cliffdelta}: A non-parametric effect size measure that quantifies the difference between two distributions. It ranges from -1 to +1, with 0 indicating no difference and ±1 indicating complete separation between the distributions. We chose Cliff's delta to robustly compare the verbatim match frequencies between $D_{old}$ and $D_{new}$, as it is less sensitive to outliers and does not assume a particular distribution shape.
    \item Kolmogorov-Smirnov (KS) distance \cite{KolmogorovSmirnovDist}: A non-parametric test that measures the maximum distance between the cumulative distribution functions of two distributions. It ranges from 0 to 1, with larger values indicating a greater difference between the distributions. We used the KS distance to complement Cliff's delta in assessing the dissimilarity between the verbatim match frequency distributions of $D_{old}$ and $D_{new}$.
    \item Kruskal-Wallis H test \cite{kruskalWallisTest}: A non-parametric test that determines whether there are statistically significant differences between the distributions of two or more groups. We employed the Kruskal-Wallis H test to assess the overall significance of the differences in verbatim match frequencies between $D_{old}$ and $D_{new}$ for each language model. This test is suitable for our analysis because it does not assume normality or homogeneity of variances, making it robust to the characteristics of our data.
\end{itemize}

By applying these statistical measures to the arrays of frequencies $F$, we aim to quantify the significance of the differences in verbatim regurgitation between the datasets that the language models were likely trained on and those that were released after the models' training cutoff dates.

The proposed methodology enables us to investigate the extent of verbatim regurgitation in language models and shed light on the influence of dataset age and availability on their ability to generate novel content. The insights gained from this study can inform the development of strategies for mitigating or detecting verbatim regurgitation in generated text and highlight the importance of considering the potential for such regurgitation when using language models for various applications.


\section{Experiments}
\begin{figure}[t!]
    \centering
    \includegraphics[width=\linewidth]{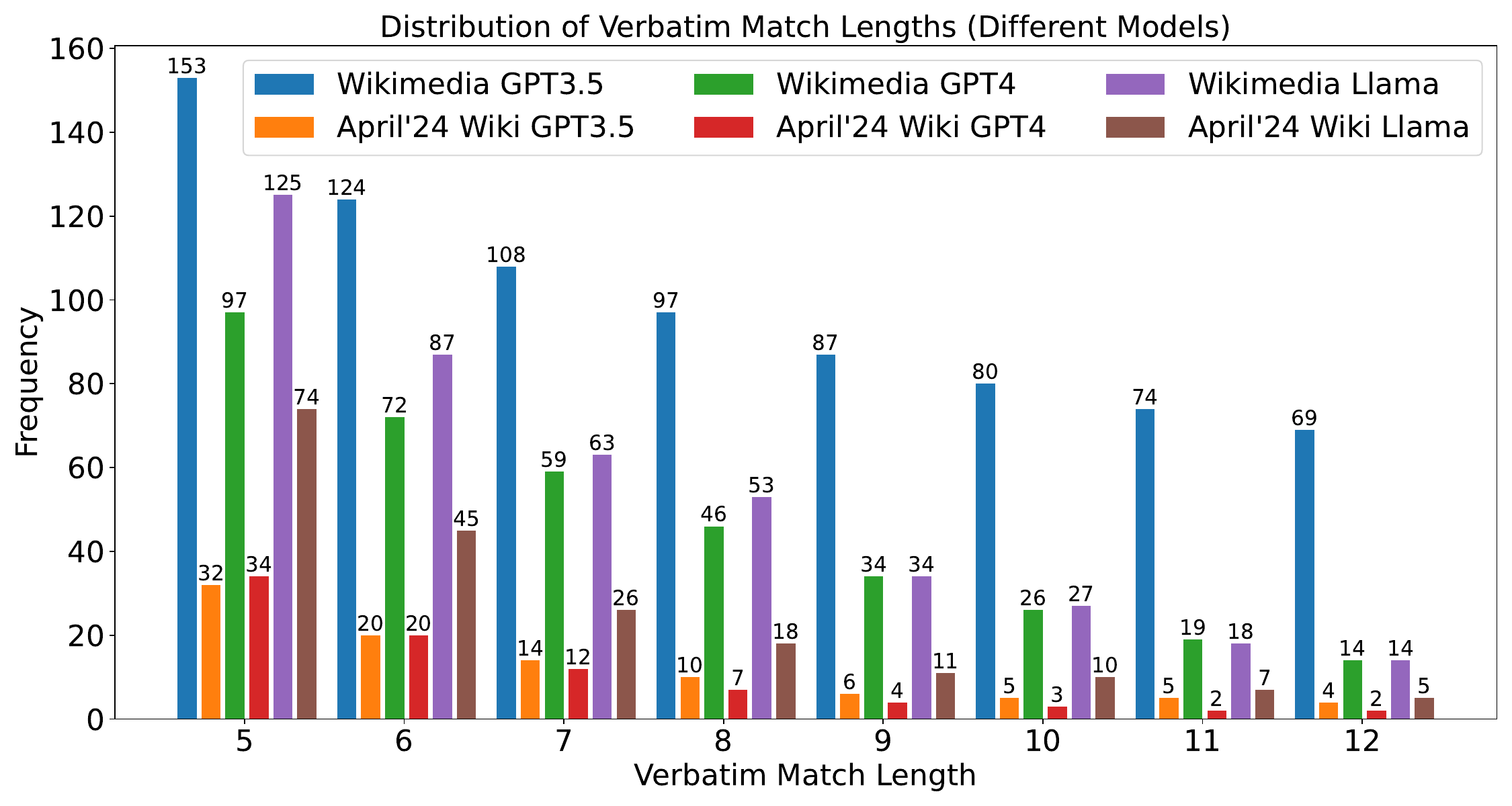}
    \caption{Distribution of verbatim match frequencies for different language models on the Wikipedia dataset. The results show a consistent pattern across all three models (GPT-3.5, GPT-4, and LLAMA), where the frequency of verbatim matches is notably higher for articles published before the training cutoff dates (Wikimedia) compared to those published after (April 2024 Wiki). This difference is particularly evident for longer verbatim match lengths, with the Wikimedia frequencies maintaining higher values even at substring lengths of 10 words or more. In contrast, the April Wiki frequencies exhibit a sharper decline as the verbatim match length increases. Statistical tests (Cliff's Delta, KS Distance, and Kruskal-Wallis H Test) confirm the significance of these differences, suggesting that the language models are more prone to reproducing verbatim content when the input text is sourced from their training data.}
    \label{fig:wiki_msr_gpt_llama}
\end{figure}

we now report on the results of a series of experiments to investigate the extent of verbatim regurgitation in text generated by different language models when prompted with text from various dataset sources. Our primary goal was to quantify the occurrence of verbatim regurgitation and understand the factors that influence this behavior, particularly focusing on the differences between datasets that the language models were likely trained on and those that were released after the models' training cutoff dates.

\subsection{Datasets and Language Models}
In our experiments, we utilized two types of dataset sources: Wikipedia articles and Open Educational Resource (OER) textbooks. For Wikipedia, we curated two subsets: $D_{\rm pre}$, containing approximately 600 articles with a length greater than 1,000 words, selected from a time period significantly before the language models' training cutoff dates, and $D_{\rm post}$, consisting of around 600 articles with a length greater than 1,000 words, crawled from the last few months, which are likely to be absent from the language models' training data. For OER textbooks, we selected $D_{\rm pre}$, comprising textbooks with publication dates before the language models' training cutoff dates, such as "OpenStax Biology 2e" \cite{bio2e}, "OpenStax Economics 2e" \cite{econ2e}, and "OpenStax Concepts of Biology" \cite{os_concepts}, and $D_{\rm post}$, including recently released textbooks, such as two new OpenStax textbooks on nutrition for nurses \cite{os_nutrition} and pharmacology, which are unlikely to be part of the language models' training data.

We evaluated the verbatim regurgitation behavior of three language models: GPT-3.5 (engine: gpt-3.5-turbo-1106), GPT-4 (engine: gpt-4-turbo-2024-04-09), and LLAMA (version: llama3-70B). For GPT-3.5 and GPT-4, we used a temperature setting of 0.1 to encourage more deterministic and less diverse outputs, which may increase the likelihood of verbatim regurgitation. LLAMA was used with greedy sampling, as it does not support adjustable temperature settings.

By carefully selecting these dataset sources and language models, we aim to create a controlled setting that allows us to compare the verbatim regurgitation behavior of language models when prompted with text they were likely exposed to during training ($D_{\rm pre}$) and text that they have not encountered before ($D_{\rm post}$). This setup enables us to investigate the influence of dataset age and availability on the extent of verbatim regurgitation and draw insights into the models' ability to generate novel content.

\begin{table}[t!]
    \centering
    \setlength{\tabcolsep}{10pt}
    \begin{tabular}{|c|c|c|c|c|c|}
        \hline
        \textbf{Datasets} & \textbf{Language Model} & \textbf{Cliff's Delta} & \textbf{KS Distance} & \textbf{KW H-statistic} & \textbf{KW p-value} \\
        \hline
        \multirow{3}{*}{Wikipedia} & GPT-3.5 & $-1.0$ & $1.0$ & $11.31$ & $0.0008$ \\
         & GPT-4 & $-0.828$ & $0.75$ & $7.77$ & $0.0053$ \\
         & LLAMA & $-0.578$ & $0.50$ & $3.78$ & $0.0519$ \\
        \hline
        \multirow{2}{*}{OER} & GPT-3.5 & $-0.984$ & $0.875$ & $11.12$ & $0.0009$ \\
         & LLAMA & $-0.828$ & $0.625$ & $7.80$ & $0.0052$ \\
        \hline
    \end{tabular}
    \caption{Statistical analysis results demonstrating a clear difference in the distribution of verbatim matches between datasets published before ($D_{\rm pre}$) and after ($D_{\rm post}$) the training cutoff dates for each language model on the Wikipedia and OER datasets. The negative Cliff's Delta effect size values indicate a higher probability of verbatim matches in $D_{\rm pre}$. The Kolmogorov-Smirnov (KS) test statistic measures the maximum distance between the cumulative distributions of $D_{\rm pre}$ and $D_{\rm post}$, with larger values suggesting a greater difference. The Kruskal-Wallis (KW) test statistic assesses the overall difference in median verbatim match frequencies across the datasets, with the corresponding p-values testing the null hypothesis of equal medians. Significant p-values (< 0.05) reject this null hypothesis. These results consistently demonstrate that verbatim match frequencies are significantly higher for $D_{\rm pre}$ compared to $D_{\rm post}$.}
    \label{tab:statistical_analysis}
\end{table}
\begin{figure}[t!]
    \centering
    \includegraphics[width=\linewidth]{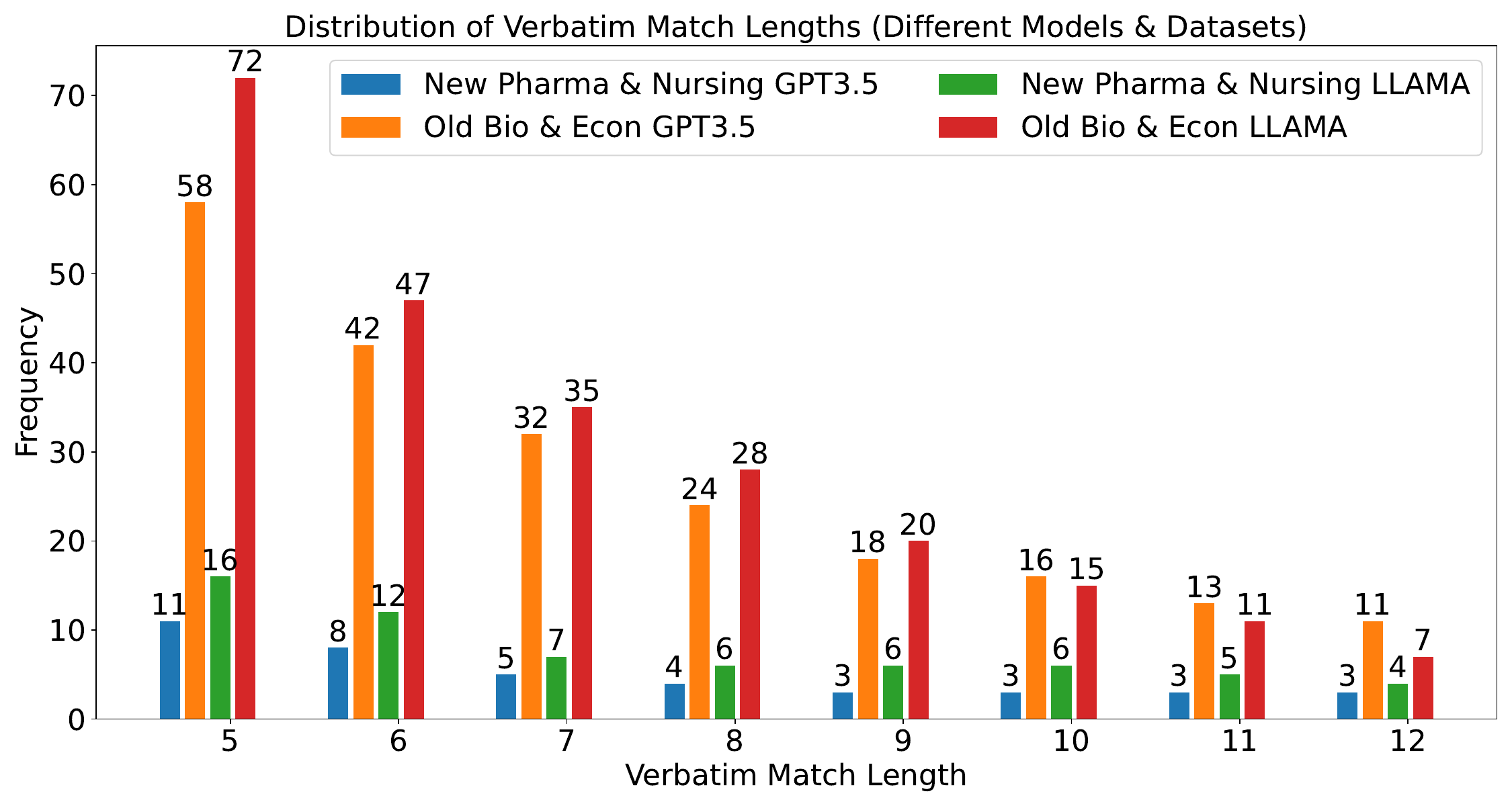}
    \caption{Distribution of verbatim match frequencies for GPT-3.5 and LLAMA on the OER textbook dataset. The frequency of verbatim matches is consistently higher for the textbooks published before the training cutoff dates (Old Bio \& Econ) compared to the recently released textbooks (New Pharma \& Nursing). Statistical tests (Cliff's Delta, KS Distance, and Kruskal-Wallis H Test) confirm the significance of these differences.}
    \label{fig:openstax_msr_gpt_llama}
\end{figure}

\subsection{Primary Experiments for Verbatim Regurgitation Analysis}
In our base experiments, we applied the Many-Shot Regurgitation (MSR) prompting technique to each combination of language model and dataset source. We used the following many shot configurations: 6 shots (3 human and 3 assistant turns). For each source text, we obtained the generated text ($T_n'$) and calculated the frequency of exact verbatim matches for substring lengths ranging from 5 to 12 words.

To quantify the differences in verbatim regurgitation between the datasets published before and after the training cutoff dates ($D_{\rm pre}$ and $D_{\rm post}$), we employed several statistical measures: Cliff's Delta, Kolmogorov-Smirnov (KS) Distance, and the Kruskal-Wallis H Test. These measures are described in detail in the Methodology section.

Figure~\ref{fig:wiki_msr_gpt_llama} presents the distribution of verbatim match frequencies for different language models on the Wikipedia dataset. The results show that the frequency of verbatim matches is consistently higher for the articles published before the training cutoff dates (Wikimedia) compared to those published after (April 2024 Wiki), across all three language models (GPT-3.5, GPT-4, and LLAMA). The statistical analysis results, summarized in Table \ref{tab:statistical_analysis}, confirm the significance of these differences, with GPT-3.5 showing the most substantial difference (Cliff's Delta $= -1.0$, KS Distance $= 1.0$, H-statistic $= 11.31$, p $= 0.0008$), followed by GPT-4 (Cliff's Delta $= -0.828$, KS Distance $= 0.75$, H-statistic $= 7.77$, p $= 0.0053$) and LLAMA (Cliff's Delta $= -0.578$, KS Distance $= 0.5$, H-statistic $= 3.78$, p $= 0.0519$).

Similarly, figure~\ref{fig:openstax_msr_gpt_llama} displays the distribution of verbatim match frequencies for GPT-3.5 and LLAMA on the OER textbook dataset. The frequency of verbatim matches is notably higher for the textbooks published before the training cutoff dates (Old Bio \& Econ) compared to the recently released textbooks (New Pharma \& Nursing). Table \ref{tab:statistical_analysis} shows that GPT-3.5 exhibits a highly significant difference (Cliff's Delta $= -0.984$, KS Distance $= 0.875$, H-statistic $= 11.12$, p $= 0.0009$), while LLAMA also shows a substantial difference (Cliff's Delta $= -0.828$, KS Distance $= 0.625$, H-statistic $= 7.80$, p $= 0.0052$).

The results of our verbatim regurgitation analysis demonstrate a clear difference in the distribution of verbatim matches between the datasets published before and after the training cutoff dates for both dataset sources. The negative Cliff's Delta values, substantial KS distances, and significant Kruskal-Wallis H-statistics consistently indicate that verbatim match frequencies are higher for $D_{\rm pre}$ compared to $D_{\rm post}$.

\begin{figure}[t!]
    \centering
    \includegraphics[width=\linewidth]{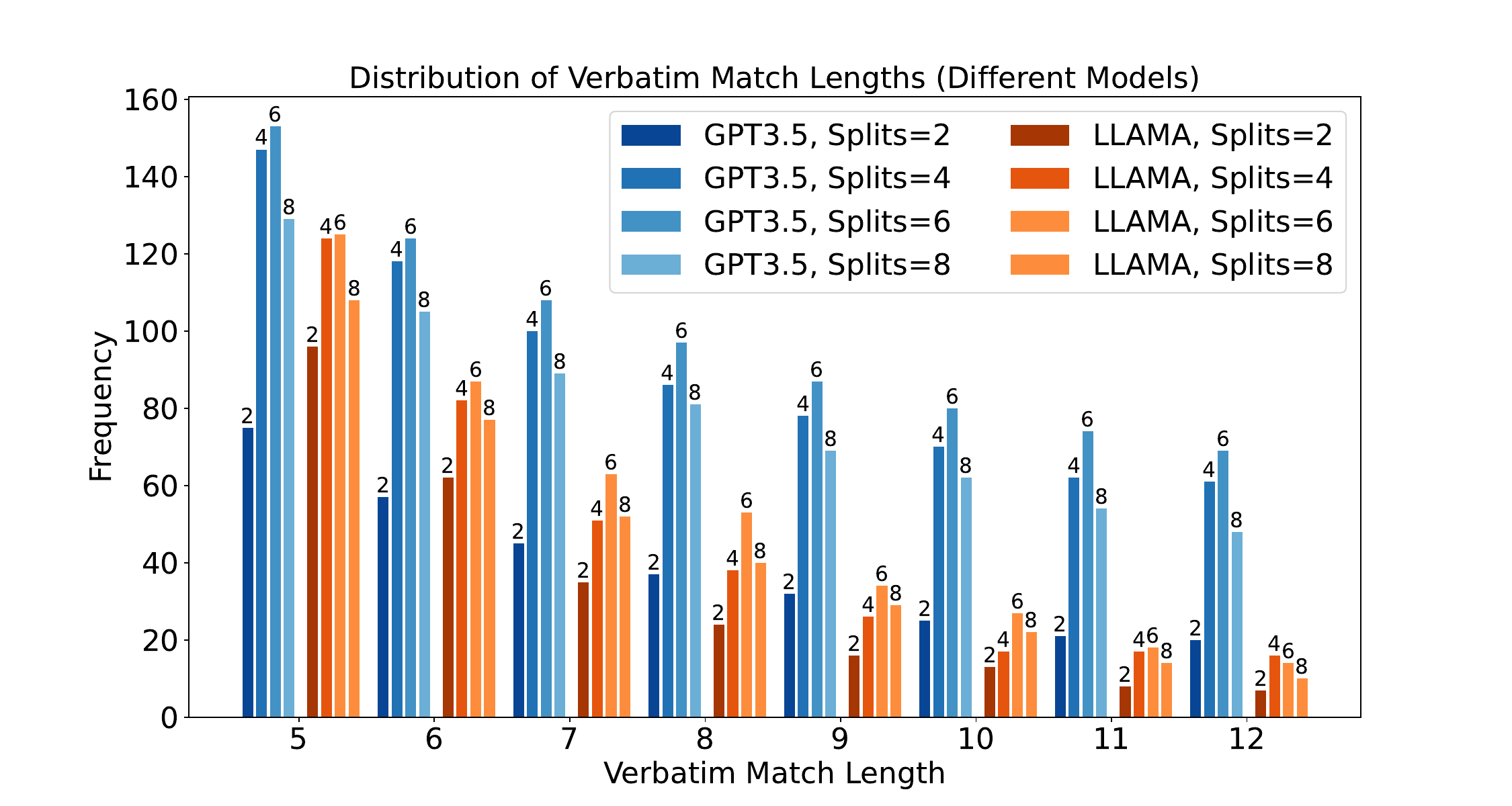}
    \caption{Effect of the number of shots on the occurrence of verbatim regurgitation for GPT-3.5 and LLAMA models on the Wikipedia dataset. The results show that increasing the number of shots generally leads to higher frequencies of verbatim matches, with 6 shots (3 human and 3 assistant turns) yielding the best results.}
    \label{fig:splits_many_shot}
\end{figure}

\subsection{Ablation Experiments}
To gain deeper insights into the factors that influence the occurrence of verbatim regurgitation, we conducted two ablation experiments: analysis of shot count and temperature control.

\subsubsection{Analysis of Shot Count}
In this experiment, we investigated the effect of the number of shots on the occurrence of verbatim regurgitation. Drawing inspiration from the many-shot jailbreaking approach, which suggests that using a higher number of shots can lead to better results, we varied the number of shots from 2 to 8 while keeping other parameters constant. We applied the MSR prompting technique to the Wikipedia articles dataset using GPT-3.5 and LLAMA models.

Figure \ref{fig:splits_many_shot} shows the distribution of verbatim match frequencies for different numbers of shots. We observed that increasing the number of shots generally leads to higher frequencies of verbatim matches for both GPT-3.5 and LLAMA models, with 6 shots (3 human and 3 assistant turns) yielding the best results. This finding suggests that dividing the source text into an appropriate number of segments is crucial for the effectiveness of the MSR prompting technique in eliciting verbatim regurgitation.

\subsubsection{Temperature Control}
In this experiment, we explored the effect of temperature settings on verbatim regurgitation. We conducted experiments with the GPT-3.5 model using two temperature values: 0.1 and 0.7. The experiments were performed on the Wikipedia dataset, and we measured the frequency of verbatim matches for substring lengths ranging from 4 to 15 words.

The results show that increasing the temperature from 0.1 to 0.7 leads to a decrease in the frequency of verbatim matches across all substring lengths. For example, at a substring length of 4 words, the frequency of verbatim matches decreases from 195 at a temperature of 0.1 to 162 at a temperature of 0.7. Similarly, for a substring length of 15 words, the frequency of verbatim matches decreases from 57 at a temperature of 0.1 to 31 at a temperature of 0.7.
To quantify the statistical significance of the differences in verbatim match frequencies between the two temperature settings, we employed Cliff's Delta, Kolmogorov-Smirnov (KS) Distance, and the Kruskal-Wallis H Test. The Cliff's Delta value of 0.5625 indicates a moderate effect size, suggesting that the verbatim match frequencies tend to be higher at a temperature of 0.1 compared to 0.7. The KS Distance of 0.5 further supports this observation, indicating a notable difference between the cumulative distribution functions of the two temperature settings.

\subsubsection{Sensitivity of Verbatim Regurgitation Detection to Input Text Length}
In this experiment, we investigated the effect of input text length on the occurrence of verbatim regurgitation. We selected 600 articles each from the Wikipedia datasets published before and after the training cutoff dates (Wikimedia and April Wiki, respectively) and truncated them to various lengths: 75, 125, 250, 500, and 1000 words. We then applied the MSR prompting technique with 6 shots (3 human and 3 assistant turns) and calculated the frequency of exact verbatim matches for substring lengths ranging from 5 to 12 words.

It is important to note that the maximum substring length that can be analyzed is limited by the ratio of the input text length ($L$) to the number of splits ($s$). The length of each split can be approximated as $L/s$, which determines the upper limit of the substring length that can be meaningfully analyzed for verbatim matches. As the input text length decreases while keeping the number of splits constant, the length of each split also decreases, reducing the maximum substring length that can be effectively analyzed.

Furthermore, as the input text length decreases, the distribution of verbatim match frequencies becomes more sparse. With shorter input texts, there are fewer opportunities for longer substring matches to occur, affecting the statistical analysis. If the distribution becomes too sparse, it may be necessary to adjust the analytical approach, such as focusing on shorter substring lengths or increasing the input text length while keeping the number of splits constant.
For each language model and dataset source combination, we obtain an array of frequencies $F = [f_{\ell_{\rm min}}, f_{\ell_{\rm min}+1}, ..., f_{l\ell_{\rm max}}]$ corresponding to verbatim match length thresholds from $l\ell_{\rm min}$ to $\ell_{\rm max}$ words.
In this experiment, $\ell_{\rm min}$ is set to 5, and $\ell_{\rm max}$ is determined by the ratio $L/s$. For example, with an input text length of 75 words and 6 splits, the average split length is approximately 12 words, so $\ell_{\rm max}$ is set to 12.

\begin{figure}[t!]
    \centering
    \includegraphics[width=\linewidth]{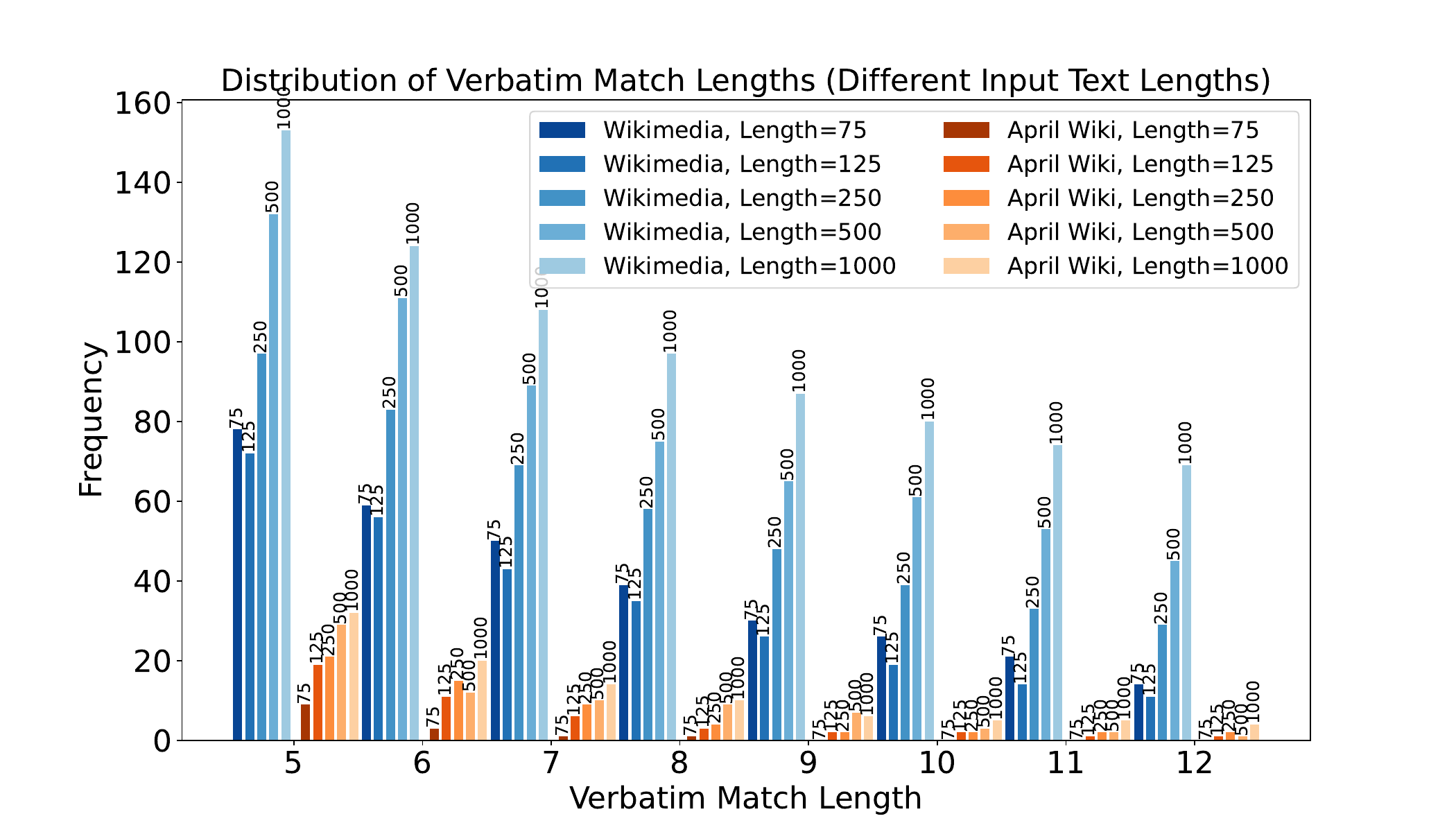}
    \caption{Distribution of verbatim match frequencies for GPT-3.5 at different input text lengths on the Wikipedia dataset. The results demonstrate that MSR prompting consistently yields higher verbatim match frequencies for articles published before the training cutoff dates (Wikimedia) compared to those published after (April 2024 Wiki), even at shorter input text lengths. However, as the input text length decreases, the maximum verbatim match length that can be meaningfully analyzed is limited by the ratio of the input text length ($L$) to the number of splits ($s$), approximated as $L/s$, resulting in a more sparse distribution of verbatim match frequencies.}
    \label{fig:length_control_gpt3_5}
\end{figure}
\begin{table}[t!]
    \centering
    \resizebox{0.9\textwidth}{!}{%
    \begin{tabular}{|c|c|c|c|c|}
        \hline
        \textbf{Input Text Length} & \textbf{Cliff's Delta} & \textbf{KS Distance} & \textbf{KW H-statistic} & \textbf{KW p-value} \\
        \hline
        $75$ & $-1.0$ & $1.0$ & $11.48$ & $0.0007$ \\
        $125$ & $-0.91$ & $0.75$ & $9.33$ & $0.0023$ \\
        $250$ & $-1.0$ & $1.0$ & $11.46$ & $0.0007$ \\
        $500$ & $-1.0$ & $1.0$ & $11.29$ & $0.0008$ \\
        $1000$ & $-1.0$ & $1.0$ & $11.31$ & $0.0008$ \\
        \hline
    \end{tabular}
    }
    \caption{Statistical analysis results comparing the distribution of verbatim match frequencies between Wikimedia and April Wiki articles at different input text lengths. The negative Cliff's Delta values, high KS Distance values, and significant Kruskal-Wallis H-statistic p-values (all $< 0.05$) consistently indicate that the Wikimedia articles have higher verbatim match frequencies than the April 2024 Wiki articles across all input text lengths. These results demonstrate the robustness of the MSR prompting technique in detecting verbatim regurgitation across a range of input text lengths, while also highlighting the limitations imposed by the relationship between the input text length and the number of splits when analyzing the distribution of verbatim match frequencies.}
    \label{tab:length_control_stats}
\end{table}

Figure \ref{fig:length_control_gpt3_5} presents the distribution of verbatim match frequencies for GPT-3.5 at different input text lengths. The results show that the MSR prompting technique consistently yields higher verbatim match frequencies for articles published before the training cutoff dates (Wikimedia) compared to those published after (April Wiki), even at shorter input text lengths.

To quantify the differences in verbatim regurgitation between the Wikimedia and April Wiki datasets at various input text lengths, we employed Cliff's Delta, Kolmogorov-Smirnov (KS) Distance, and the Kruskal-Wallis H Test. Table \ref{tab:length_control_stats} summarizes the statistical analysis results for each input text length.

The results demonstrate that the MSR prompting technique is effective in eliciting verbatim regurgitation even at shorter input text lengths. The Cliff's Delta values of -1.0 or close to -1.0, along with the large KS distances and significant Kruskal-Wallis H-statistics (p < 0.05), indicate that the verbatim match frequencies are consistently higher for the Wikimedia dataset compared to the April Wiki dataset across all input text lengths.

These findings suggest that the effectiveness of the MSR prompting technique in detecting verbatim regurgitation is robust to variations in input text length, at least down to a length of 75 words. This highlights the potential of the MSR prompting technique as a tool for identifying instances of verbatim regurgitation in language models, even when working with shorter input texts.

However, it is essential to consider the limitations imposed by the relationship between the input text length and the number of splits when analyzing the distribution of verbatim match frequencies. Future research could explore the impact of varying the number of splits in conjunction with the input text length to further investigate the boundaries and optimal configurations for detecting verbatim regurgitation using the MSR prompting technique.

\section{Conclusion}
In this paper, we have introduced Many-Shot Regurgitation (MSR) prompting, a novel approach for investigating verbatim content reproduction in large language models. MSR represents a new type of membership inference attack (MIA) that operates in a black-box setting, requiring only the ability to prompt the model and observe its outputs. We quantified the occurrence of verbatim matches using the Longest Common Substring algorithm and devised a novel evaluation strategy that compared the distribution of verbatim matches between two carefully curated dataset sources: Wikipedia articles and Open Educational Resource textbooks. The key innovation in our evaluation approach was the inclusion of documents published after the LLMs' training cutoff dates ($D_{\rm post}$), serving as a control group that the models could not have encountered during training.
Our experiments to quantify the extent of verbatim regurgitation in state-of-the-art LLMs such as GPT-3.5, GPT-4, and LLAMA revealed a striking difference in the distribution of verbatim matches between datasets that the models were likely exposed to during training ($D_{\rm pre}$) and those published after the models' training cutoff dates ($D_{\rm post}$). Robust statistical measures, including Cliff's delta, Kolmogorov-Smirnov distance, and the Kruskal-Wallis H test, consistently indicated significantly higher verbatim match frequencies for $D_{\rm pre}$ compared to $D_{\rm post}$. Furthermore, our ablation experiments shed light on the factors affecting the effectiveness of the MSR technique, such as the number of shots and the temperature settings of LLMs. In sum, our work highlights the effectiveness of MSR prompting as a new MIA framework for studying LLMs in real-world settings where access to model internals is limited.

\section*{Acknowledgments}
This work was supported by NSF grant 1842378, ONR grant N0014-20-1-2534, AFOSR grant FA9550-22-1-0060, a Vannevar Bush Faculty Fellowship, OpenAI, and ONR grant N00014-18-1-2047.

\bibliographystyle{unsrt}  
\bibliography{mybibliography}

\end{document}